\documentclass[preprint, 12pt]{amsart}
\usepackage[table]{xcolor}
\usepackage{times}
\usepackage{amssymb, amsmath}
\usepackage{amsthm}
\usepackage{tikz}
\usepackage{bm}
\usetikzlibrary{matrix,arrows}
\usepackage{enumitem}
\usepackage{booktabs}
\usepackage[section]{placeins}
\usepackage{color}
\usepackage{multirow}

\theoremstyle{plain}

\theoremstyle{remark}

\definecolor{Gray}{gray}{0.95}
\newcolumntype{g}{>{\columncolor{Gray}}c}

\graphicspath{{./images/}}

\begin{document}
\title[ML forecasting in FX markets]{Machine learning based forecasting of significant daily returns in foreign exchange markets}
\author{Firuz Kamalov$^1$$^{*}$ and  Ikhlaas Gurrib$^2$}

\address{$^{1}$ Canadian University Dubai, Dubai, UAE.}
\email{\textcolor[rgb]{0.00,0.00,0.84}{firuz@cud.ac.ae}}

\address{$^{2}$ Canadian University Dubai, Dubai, UAE.}
\email{\textcolor[rgb]{0.00,0.00,0.84}{ikhlaas@cud.ac.ae}}

%\date{\today}
\date{\today
\newline \indent $^{*}$ Corresponding author}

\begin{abstract}
Asset value forecasting has always attracted an enormous amount of interest among researchers in quantitative analysis. The advent of modern machine learning models has introduced new tools to tackle this classical problem. 
In this paper, we  apply machine learning algorithms to hitherto unexplored question of forecasting instances of \textit{significant} fluctuations in currency exchange rates. We perform analysis of nine modern machine learning algorithms using data on four major currency pairs over a 10 year period. A key contribution is the novel use of outlier detection methods for this purpose. Numerical experiments show that outlier detection methods substantially outperform traditional machine learning and  finance techniques. In addition, we show that a recently proposed new outlier detection method PKDE produces  best overall results. Our findings hold across different currency pairs, significance levels, and time horizons indicating the robustness of the proposed method.
\end{abstract}

\maketitle

%-----------------------------------------------------------------------------------------------------------------------------------------------------
%-----------------------------------------------------------------------------------------------------------------------------------------------------
\section{Introduction}
Forecasting the future value of financial assets is perhaps the most important question in finance. The implications of being able to correctly determine the future value of a financial asset are hard to underestimate. Therefore, an enormous amount of research has been dedicated to the topic. A plethora of rules, metrics, and indicators exist that attempt to answer this age old question. Until recently most of the effort has been based on using financial indicators \cite{gurrib1}. However, economic and financial data contain a large amount of information with many complex inner relationships. It would be nearly impossible to build a predictive model that captures all the relationships. With the current advances in technology and increased availability of data researchers have started applying machine learning tools to tackle this problem  \cite{deprado, galeshchuk, gu, guresen}. Our goal is to apply modern machine learning tools to forecast significant changes in currency exchange rate. In particular, we attempt to forecast significant fluctuations in daily exchange rate based on the changes in exchange rate during the previous $p$ days.
Note that forecasting of the actual change in exchange rate is a nearly impossible task \cite{meese}. The Efficient Market Hypothesis stipulates that asset prices fully reflect all the available information and it is impossible to predict the market on a consistent basis. This observation is illustrated in Figure \ref{all_returns}, where we can see the chaotic and unpredictable nature of changes in exchange rates. Therefore, we focus our efforts on forecasting only the significant changes in exchange rate.   We believe that it is a more reasonable task since irrational market exuberance can lead to occasional overselling (or underselling) of an asset leading to an eventual significant change in its price. We attempt to learn the particular conditions that lead up to a significant change in asset price using the modern machine learning tools.
The contribution of our paper is two-fold: 

\begin{enumerate}
\item Analysis of efficacy of modern machine learning tools in forecasting  \emph{significant} fluctuations in currency exchange rates. We apply six modern machine learning algorithms to 10-year data on four major currency pairs. The results are benchmarked by the performance of an RSI model.
\item Novel application of outlier detection techniques in predicting significant daily returns. We test three outlier detection methods including a recently proposed PKDE method that produces the best overall results.
\end{enumerate}

Market timing models can be traced back to Treynor and Mazuy \cite{treynor}, where market players have ability to change their exposures based on future market movement expectations.  There is a slew of statistical finance approaches to modeling time series including moving average, exponential smoothing, ARIMA and others, in which investors conduct either technical analysis, fundamental analysis, or a combination of both. While opposing views exist as to the success of technical analysis trading in foreign exchange markets, the survey in \cite{menkhoff1} found substantial evidence of excess return when using technical indicators.  Although technical analysis has been used extensively in the literature existing studies often lack in three common aspects.  First, short coverage periods may exclude the effects of major financial events like the 2008 financial crisis.
Second, a variety of technical analysis techniques is used without robust testing other innovative tools such as machine learning.  
Third, only one exchange rate are considered, thereby limiting the generalization of their results to other exchange rate markets \cite{schulmeister}.  
Our study addresses the above issues by conducting an analysis over a sufficiently broad number of years, comparing one of the mostly used indicators, the Relative Strength Index (RSI), with machine learning techniques, and pursuing analysis over some of the most actively traded US based currency pairs.

To tackle the question of forecasting significant changes in exchange rates we can apply both supervised and unsupervised machine learning approaches. Supervised algorithms can be further implemented  as either continuous or discrete with respect to the target variable. In a continuous supervised algorithm the target variable, i.e. the forecast daily change, is assumed to be continuous-valued. Continuous supervised algorithms are often referred to as regression models. There exist a number of continuous supervised algorithms available for our purpose including least squares regression, support vector regression, and neural network regression. Similarly, in a discrete supervised algorithm the target variable is assumed to be discrete-valued.  Discrete supervised algorithms are referred to as classification algorithms. In our paper, we will use random forests, support vector classifier, and neural network classifier.  Since daily changes are continuous-valued we must first discretize them before applying classification algorithms. This is accomplished through thresholding whereby return values above the threshold will be labeled as $1$ and below the threshold as $0$. 

The problem of forecasting extreme daily changes can also be  taken on as an unsupervised learning task. In particular, we propose to approach the question at hand as an outlier detection problem. In fact, the novel and effective application of outlier detection methods to predict significant fluctuations in exchange rate is one of the key contributions of the present paper. Outlier detection algorithms analyze unlabeled data in order to determine instances that differ significantly from the rest of the data. As such it is natural to view the task of determining extreme forecast values as an outlier detection problem. There exist a number of algorithms for outlier detection including robust covariance, local outlier factor, and principal kernel density estimate that can be used for our purpose.

In our study, we carry out extensive and rigorous experiments using nine machine learning algorithms using 10-year daily historical data on four major currency exchange pairs. The machine leaning models' performance is benchmarked against an RSI based forecasting model. We use $F_1$ and recall scores to evaluate the performance of the learning algorithms. Numerical experiments  demonstrate that outlier detection methods outperform  other supervised machine learning techniques as well as the traditional financial methods. The $F_1$-scores decrease with increase in significance level while the recall scores increase. We discern that the models improve at identifying significant daily returns at higher thresholds but they fail to effectively distinguish between significant and normal returns.
The results hold across different exchange rate pairs, significance thresholds and time periods indicating the robustness of the proposed approach. 
Our paper is organized as follows. In Section 2, we present the current literature related to asset value forecasting. In Section 3, we briefly outline the machine learning and finance algorithms used in the paper. Section 4 contains the results of numerical experiments and Section 5 concludes the paper.
%-----------------------------------------------------------------------------------------------------------------------------------------------------
%-----------------------------------------------------------------------------------------------------------------------------------------------------
\section{Literature}
We broadly divide our review of the literature into two parts: finance and machine learning. The first part sets up the necessary background of existing technical analysis approaches used in finance. The  second part discusses the machine learning studies  related to financial forecasting.

\subsection{Finance}
Based on a survey of chief foreign exchange dealers in the UK the authors in \cite{taylor} found that 90\% of respondents placed some weight on technical analysis, with greater reliance on technical analysis than fundamental analysis. Similarly, after conducting a survey of foreign dealers in Hong Kong the authors in \cite{lui} found that technical analysis is considered slightly more useful in forecasting trends than fundamental analysis, and significantly more useful in predicting turning points. As suggested in \cite{epley} one of the main reasons investors use technical analysis tools is due to their reluctance to make adjustments by staying close to their anchors. The authors  confirm that adjustment to other techniques requires more effort.    The authors in \cite{gurrib1} look at market timing tools such as double crossover strategies and find superior performance compared to the naïve buy and hold strategy. The authors in \cite{pruitt} propose a trading system based on volume, RSI, and moving average that also outperformed the market after adjusting for transactions costs.  The authors  in \cite{menkhoff2} discovered that most fund managers in five countries under consideration use technical analysis. In \cite{wong} the authors  found the use of RSI and moving average to yield significant positive returns in the Singapore Stock Exchange.  It should be noted that the widely used techniques are subject to various assumptions.  For instance, in \cite{neely} the authors find that both market conditions and profitability upon using technical analysis techniques change over time.  As laid out in \cite{gurrib2}, RSI does not take into account the underlying price of a security so that one cannot interpret the same value of RSI for two differently priced securities in the same manner.  A high value of RSI for a low priced security tends to give a stronger signal to sell than for a higher priced security.  RSI can be oversensitive to relatively minor price movements as in the case of stable stocks.  Finally, RSI can behave asymmetrically in the lower versus higher ends of its spectrum, such that the derivative of the relative strength index function can show a rapid increase for small relative strength values near zero, and increases slower for larger values.  For the purpose of this study, we assume that the most actively trading currencies behave more or less similarly relative to the US dollar, and witness similar reactions to major events in financial markets when it comes to their price changes.

Since Goodman in \cite{goodman}  surveyed foreign exchange (FX) forecasting techniques among FX service providers, and found technically-oriented FX services to outperform economic-oriented services, there has been abundant research work in international finance focusing on exchange rates.  Evans in \cite{evans} finds that interest rate information contributed to more than 80\% of variation in intraday shocks of the EUR/USD.  Others focused on  forecasting volatility in foreign exchange markets, by using forecasting tools such as implied standard deviation, autoregressive and GARCH models.  As in many other prior works, out of sample forecasting in these studies was mostly based on statistical accuracy criteria like minimizing the root mean squared errors or mean direction accuracy. The use of directional change (DC) has also become popular in the literature. The authors in \cite{glattfelder} found that a DC threshold of $\theta$ is commonly followed by an overshoot (OS) with the same threshold, and that if on average a DC takes $t$ amount of time to complete, an OS takes $2t$ of physical time.  To determine the optimal threshold level for DC 
the authors in \cite{kampouridis} adjusted the DC model with genetic algorithm to optimize threshold related factors. They found the upgraded model to be superior in generating returns, compared to other techniques such as technical analysis and buy and hold.

%-----------------------------------------------------------------------------------------------------------------------------------------------------
%------------------------------------------------------------------------------------------------------------------------------------------------

\subsection{Machine Learning}
Exchange rate forecasting is a difficult task. In fact, the seminal paper by Meese and Rogoff \cite{meese} showed that traditional econometric models did not perform any better than a random walk model. The success of forecast models depends on a number of factors. In search of novel approaches researchers drew inspiration from such esoteric fields as macromolecules \cite{contreras} and behavioral science \cite{degrauwe}. In recent years machine learning has also become a popular tool for exchange rate prediction.  In \cite{bissoondeeal}, the authors compared artificial neural networks (ANN) to conventional models in the context of exchange rate forecast and discovered that neural networks produce better results. The authors in  \cite{onder} showed that due to their flexibility ANNs outperform conventional time-series models in estimating macroeconomic parameters. However, ANN flexibility may be a double edged sword in that it can overfit the data. In \cite{thinyane}, the authors applied genetic algorithms and ANNs to exchange rates and showed that ANNs can produce high profits on training sets but underperform on the test set. Therefore, a simpler network architecture is preferred to avoid overfitting. The authors in \cite{galeshchuk}, applied artificial neural networks using a 5-10-1 architecture  to predict exchange rate between three major currency pairs. They used daily, monthly, and quarterly time steps in their study. The results showed that short term predictions are more accurate than longer term forecast. More advanced neural network architectures such as LSTMs have been recently deployed to forecast stock prices \cite{kamalov3}. LSTM have the ability to process sequential data that is ideally suited in time series prediction.

Support vector machines are powerful learning algorithms. They have been applied in various contexts including exchange rate forecasting. To account for non-linearities in time series the authors in\cite{lin} propose a hybrid forecasting model that combines empirical mode decomposition and support vector regression. Test results based on three currency exchange pairs show that the hybrid model outperforms several individual models in exchange rate forecasting. The authors in \cite{plakan}, applied a number of machine learning algorithms including SVM to predict the direction of exchange rate based on market sentiment. The findings indicate that investor sentiment as expressed on public message boards can be useful in market prediction.
Random forest is a classic machine learning algorithm. It is an efficient technique that has been used in different forecasting problems including energy \cite{wang2}, remote sensing \cite{belgiu} and others.

Outlier detection is an unsupervised machine learning algorithm that is used to identify anomaly instances. They have been employed in diverse fields such as including biology, security, astronomy many others \cite{campos, kamalov}. 
In finance, they have been used for fraud detection, risk modeling, customer behavior analytics and other applications \cite{paula, phua, wang}. 
In \cite{nian}, the authors propose spectral ranking of anomalousness based on the first nonprincipal eigenvector of the Laplacian matrix. The approach is successfully applied to auto insurance claim data to detect instances of fraud. 
The authors in \cite {ram} use density estimation trees to analyze the distribution of data points and calculate their 
outlier scores. The authors apply their approach to detect fraudulent financial transactions. 
Despite a wide array of applications outlier detection has not been, to the best of our knowledge, used in forecasting exchange rates.

%-----------------------------------------------------------------------------------------------------------------------------------------------------
%-----------------------------------------------------------------------------------------------------------------------------------------------------
\section{Forecasting methods}

In this section, we give a brief outline of the forecasting methods employed in the paper. We can broadly divide the methods used in our study into 3 categories: financial, supervised, and unsupervised methods. Financial methods refer to the traditional indicators such as the relative strength index used by finance practitioners. Supervised methods refer to machine learning algorithms that are based on a known value of the target variable. Similarly, unsupervised algorithms refer to learning algorithms where the value of target variable is unknown.

\subsection{Financial methods}

One of the most popular indicators used in investment analysis to signal an imminent reversal in the price movement of a security is the RSI.  It measures the cumulative gain relative to cumulative loss in asset price. The RSI is calculated as follows:
\begin{equation}
RSI = 100-\frac{100}{1-RS},
\end{equation}
where RS is the ratio of sum of gains to sum of losses over a period. A look-back period of 14 days is often used as the default setting.  The RSI consists of predominantly two boundary levels: 30 and 70.  These levels are initially set as default by trading systems, but can be adjusted as the trader or investor sees fit.  These levels capture trading signals that can be acted upon by the traders.  An index crossing above 70 is indicative of an overbought situation that is expected to lead to a reversal in subsequent prices. Likewise, an index crossing below 30 is indicative of an oversold situation with an expectation of future price increase.

%-----------------------------------------------------------------------------------------------------------------------------------------------------
\subsection{Supervised methods}
Supervised learning models are used on data with known target values. Concretely, given a set of input features (variables) $\boldsymbol{X}$ and output (target) values $y$ a supervised learning algorithm builds a predictive model to best estimate the value of $y$ based on the values of $\boldsymbol{X}$. Supervised models can be categorized into 2 groups: continuous and discrete. 

%-----------------------------------------------------------------------------------------------------------------------------------------------------
\subsubsection{Discrete models}
Discrete supervised models, called classifiers, are applied to data with a discrete-valued target variable. The target variable can be binary as in the case of patient being sick (yes/no) or multilabel as in the case of handwriting recognition of single digits (0, 1, 2, ..9). We note that in our case the target variable - daily return of currency exchange rate - is continuous-valued. Therefore, the target variable must be discretized prior to applying a classifier. In this paper, we employ 3 classifiers: random forests (RF), support vector classifier (SVC), and neural network classifier (NNC).

\begin{itemize}
\item RF is an ensemble classifier that aggregates a collection of decision trees to produce a prediction \cite{breiman}. It is an efficient, robust classifier characterized by low variance.  Decision tree is a classifier that has a tree-like structure. 
%(Figure \ref{dt_illust}). 
It is constructed by splitting data along the features based on information gain. Each node in a decision tree corresponds to a feature split and the branches correspond to the relative value of a data point at the node feature. RF generates new trees by repeatedly sampling the data and fitting a tree on the sampled data. As a result RF reduces overfitting.  The output of RF is based on the mode of all decision tree outputs. 
%\begin{figure}[h!]
%\center
%\includegraphics[scale=0.5]{credit_dt.png}
%\caption{Sample decision tree}
%\label{dt_illust}
%\end{figure}

\item SVC is a powerful non-linear classifier that has proven to be successful in various fields such as hand-written character recognition and protein classification \cite{sun, zhang}. SVC was originally designed as a large margin linear classifier, i.e., it builds a hyperplane with maximum distance between two classes (Figure \ref{svm_illust}). However, its popularity is due to the kernel trick that allows it to use a nonlinear function such as a Gaussian to build nonlinear boundaries between classes. 

\begin{figure}[h!]
\center
\includegraphics[scale=0.6]{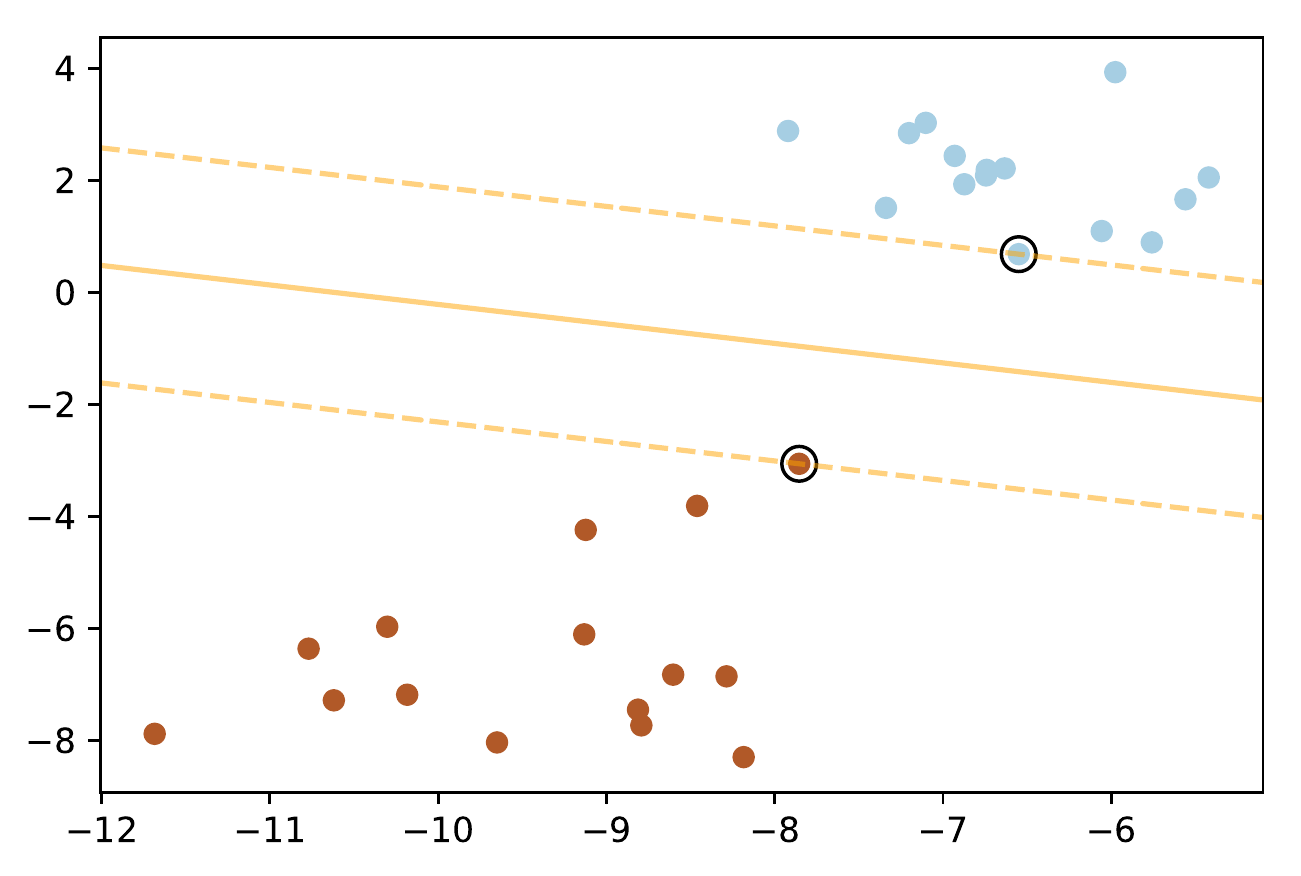}
\caption{SVC finds the maximum margin separating hyperplane.}
\label{svm_illust}
\end{figure}

\newpage

\item NNC is a flexible nonlinear classifier that makes it possible to customize the learning model to a particular problem. NNC has been used successfully in a number of applications including image and speech recognition \cite{dong, xiong}. The design of a neural network is inspired by the brain structure where the neurons are interconnected by synapses. A fully connected network consists of an input layer, hidden layer(s), and the output layer (Figure \ref{ann_illust}). The hidden layers allow the network to learn new features to classify the data. 

\begin{figure}[h!]
\center
\includegraphics[scale=0.6]{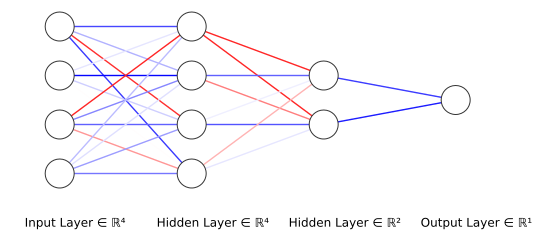}
\caption{ANN with two hidden layers.}
\label{ann_illust}
\end{figure}
\end{itemize}

%-----------------------------------------------------------------------------------------------------------------------------------------------------
\subsubsection{Continuous models}
Continuous supervised algorithms, called regression models, assume that the target variable is continuous-valued. In our paper, we employ 3 regression models to forecast exchange rate returns: least squares regression (OLS), support vector regression (SVR), and neural network regression (NNR).

\begin{itemize}
\item OLS is a traditional tool in statistics that is used for linear approximation of $y$ based on $\boldsymbol{X}$. Geometrically, OLS determines the equation of the hyperplane that best fits the given data. The best fit is accomplished by minimizing the mean squared error between the true and predicted values of $y$. 

\item SVR is similar to SVM in that it uses only a portion of the data. SVR builds a model ignoring the data points which are close to the hyperplane. The kernel trick allows SVR to easily create nonlinear models.  

\item NNR is a powerful regression model that is based on neural network architecture. Its main difference with NNC is the loss function. While NNC uses binary or categorical-crossentropy, NNR uses mean squared error as its loss function. The flexibility of the neural network architecture allows user to customize the model to the specific problem.
\end{itemize}

%-----------------------------------------------------------------------------------------------------------------------------------------------------
\subsection{Unsupervised methods}
\begin{figure}[h!]
\center
\includegraphics[scale=0.6]{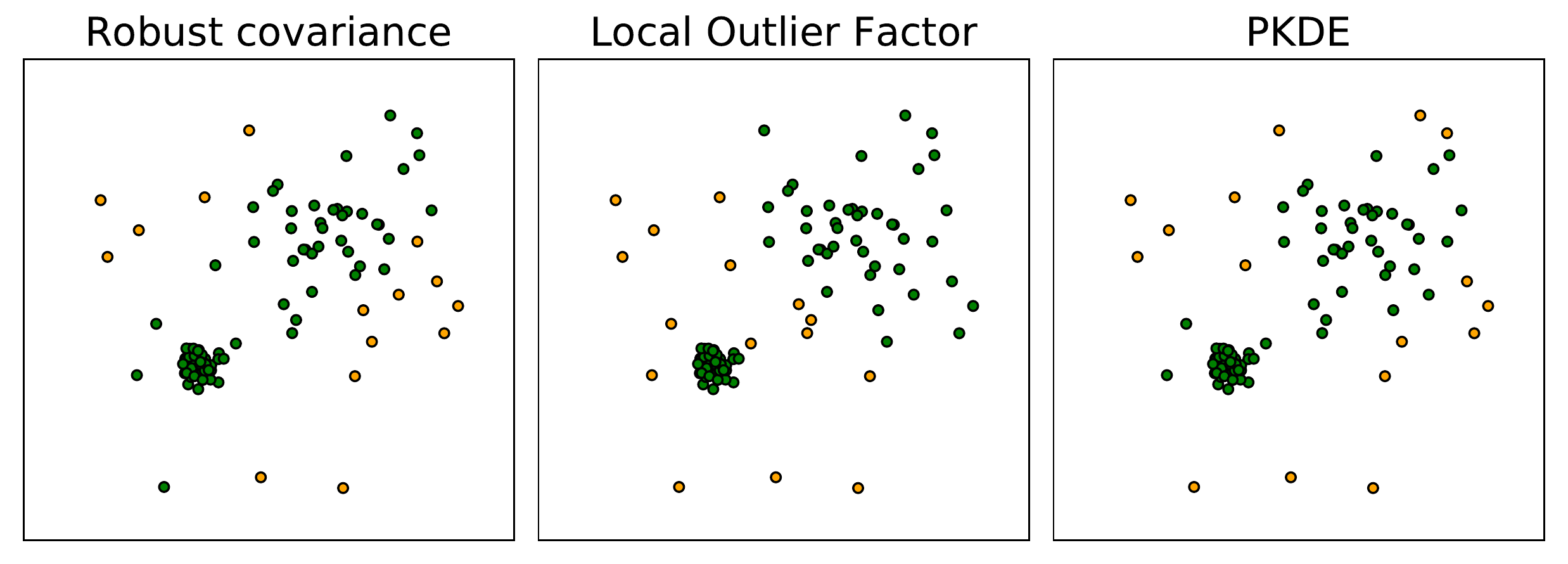}
\caption{Outlier detection methods. Red colored points indicate outliers and blue normal points.}
\label{outliers}
\end{figure}

Unsupervised machine learning algorithms are designed for tasks where the target variable does not exist or its values are unknown. There exist different families of unsupervised algorithms. In this paper, we focus on outlier detection algorithms. Since we consider significant fluctuations in exchange rate as abnormal events it seems natural to apply outlier detection methods to detect the said fluctuations. In our study, we employ three outlier detection algorithms namely robust covariance (RC), local outlier factor (LOF), and principal kernel density estimate (PKDE) (Figure \ref{outliers}).

\begin{itemize}
\item RC algorithm assumes that  normal data follows a fixed distribution such as Gaussian. Then the parameters of the distribution are calculated based on the data \cite{rousseeuw}. The points in the data that have the lowest probability based on the assumed distribution are considered as outliers. RC is a simple and efficient algorithm, and often shows good performance. However, as a global method it includes potential outliers in the calculation of the distribution parameters and therefore may be affected by exogenous instances. It may also fail to  detect outliers in variable density populations. Nevertheless, its efficiency and generally accurate performance make it a good candidate algorithm.

\item LOF is designed to address the issue of variable density. If a population consists of clusters with different densities then global methods such as robust covariance may fail to detect some of the outliers. LOF calculates the local density at the given point and compares it to the density at neighboring points. The idea is to detect points that have significantly different densities than their neighbors \cite{breunig}. LOF performs well in many different scenarios. However, it may be computationally burdensome as it requires determining neighbor points. 

\item PKDE is an outlier detection method designed to perform well in high dimensional data \cite{kamalov}. PKDE works by first reducing the dimension of the data using principal component analysis (PCA) and then applying kernel density estimation (KDE) to measure the outlyingness of a point. PCA is a common dimensionality reduction tool which makes it possible to reduce the dimensions of the data with little loss of information. PCA is performed by sequentially finding the orthogonal directions in which the data has maximum variance and then projecting the original data onto the subspace of sufficient total variance. In practice, PCA is calculated through singular value decomposition of the data covariance matrix. After the dimensonality reduction we use KDE to estimate the likelihood of a given point being an outlier. KDE is a powerful nonparametric  probability density estimation technique that has had a wide array of applications \cite{kamalov2, kamalov4, gurrib3, silverman}. The outlyingness likelihood is calculated as the average distance between a given point and the rest of the data where the distance is measured via a kernel function. The kernel used in KDE is usually a nonlinear function such as a Gaussian. Thus the distance between two points becomes nonlinear instead of the usual Euclidean distance. Since KDE is a nonparametric approach it offers a greater flexibility than traditional parametric methods.
\end{itemize}

%-----------------------------------------------------------------------------------------------------------------------------------------------------
%-----------------------------------------------------------------------------------------------------------------------------------------------------
\section{Numerical experiments}
In this section, we present the results of our numerical experiments to forecast significant fluctuations in exchange rate as measured by daily returns. As will be evident from the following discussion outlier detection methods outperform the traditional machine learning and financial algorithms. In particular, the recently proposed PKDE method produces the most optimal results among all the considered methods. The results remain consistent across different currency pairs considered as well as varying time horizons and significance levels.

\subsection{Methodology}
\begin{figure}[h!]
\center
\includegraphics[scale=0.6]{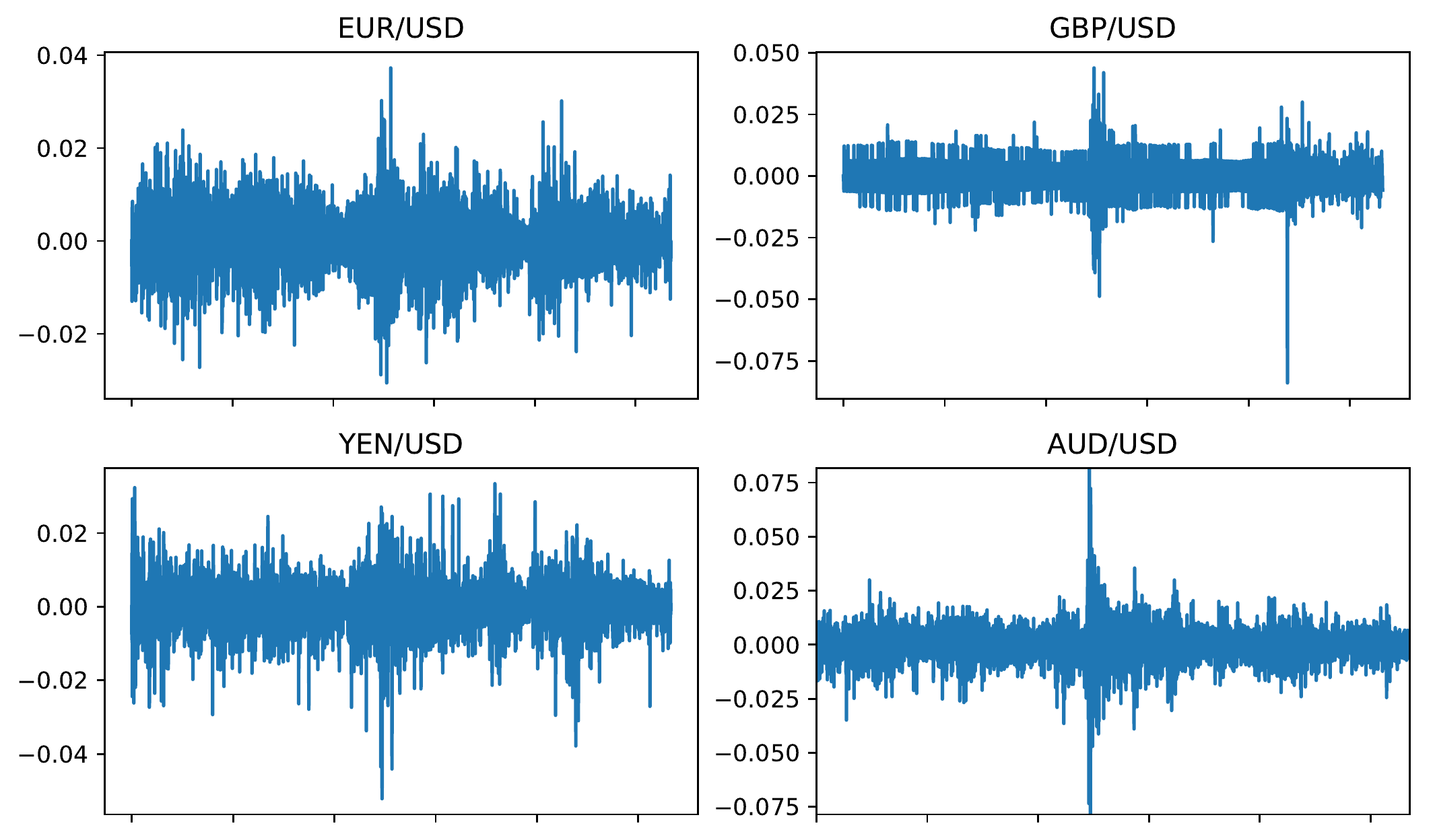}
\caption{Daily returns of currency exchange rates from Jan 1, 1999 to Sep 1, 2019.}
\label{all_returns}
\end{figure}

We use  daily exchange rate of 4 major currency pairs - USD/EUR, USD/GBP, USD/YEN, and USD/AUD - to carry out our experiments (Figure \ref{all_returns}).  The exchange rate data is taken from Jan 1, 1999 to Sep 1, 2019 \cite{macrotrends}.

\begin{figure}[h!]
\center
\includegraphics[scale=0.7]{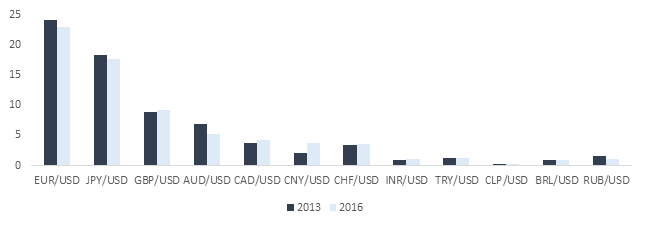}
\caption{Foreign exchange market turnover of USD based currency pairs. The data was compiled from BIS (2016).
}
\label{bar_graph}
\end{figure}

BIS (2016) reported the US dollar to stay as the dominant trading currency with involvement in 88\% of all trades.  As shown in Figure \ref{bar_graph}, the seven most actively traded currency pairs based on over the counter (OTC) transactions, were found to involve the USD on one side of the currency pair.  These include the EUR/USD, JPY/USD, GBP/USD, AUD/USD, CAD/USD, CNY/USD, and the CHF/USD.  This is in line with BIS (2016) which reported that the top five most active currencies during 2013 and 2016 were the USD, EUR, JPY, GBP and the AUD.  The USD shared 87 and 87.6 per cent of all OTC foreign exchange transactions during 2013 and 2016. 

A common way to measure change in the financial industry is using \emph{daily return}. To calculate the daily return of exchange rates we use the logarithmic ratio of two consecutive days:

\begin{equation}
\label{return_eq}
r_t = \ln \big(\frac{E_t}{E_{t-1}}\big),
\end{equation}
where $r_t$ is the return and $E_t$ is the exchange rate on day $t$. To predict the change on day $t$ we use the return values of previous $p$ days, i.e,  $\{r_{t-1}, r_{t-2},..., r_{t-p}\}$. The predictive model can thus be described via equation:

\begin{equation}
\label{model_eq}
y_t = F(r_{t-1}, r_{t-2},..., r_{t-p}),
\end{equation}
where $y_t$ is a binary variable that is $1 $ if daily return on day $t$ is significant and $0$ otherwise, and $F$ is the predictive model. The goal is to determine the optimal model $F$ that can accurately predict the value of $y_t$. To this end, we consider a number of machine learning approaches including regression, classification, and outlier detection.

In our study, we consider models based on different values of $p$: 7 days, 14 days, 30 days, and 60 days. Thus, a value of $p=30$ would imply that the model predicts the return on day $t$ using the return values of previous 30 days. We define the daily return to be significant based on a predetermined threshold  defined as a multiple of the standard deviation of daily returns.
For instance, let $\sigma$ be the standard deviation of daily returns of the USD/EUR pair over the entire period of Jan 1, 1999 to Sep 1, 2019. Then a threshold value of $1.5\sigma$ implies that any daily return value above $1.5\sigma$ or below $-1.5\sigma$ is considered significant. 
To broaden our analysis we test over a range of threshold values.

In this study, we analyze the performance of nine machine learning algorithms together with one financial method. The list of the methods is given from Table \ref{algo_list} with further details supplied in Section 3. 
\begin{table}[h!]
\centering
\caption{Machine learning and financial algorithms used in the study.}
\label{algo_list}
\begin{tabular}{@{}llll@{}}
\toprule
\textbf{Regression} & \textbf{Classification} & \begin{tabular}[c]{@{}l@{}}\textbf{Outlier}\\ \textbf{Detection}\end{tabular} & \textbf{Financial}            \\ \midrule
OLS        & RF             & RC                                                          & \multirow{3}{*}{RSI} \\
SVR        & SVC            & LOF                                                         &                      \\
NNR        & NNC            & PKDE                                                        &                      \\ \cmidrule(r){1-4}
\end{tabular}
\end{table}
The machine learning algorithms can be divided into three categories: regression, classification, and outlier detection. The RSI method is used as a benchmark. As can be seen in Table \ref{algo_list}, our analysis includes a wide variety of machine learning algorithms providing us with a  broad overview of efficacy of machine learning tools in predicting significant daily returns. 

The choice of the performance metric requires careful consideration. Since the number of significant daily returns is substantially smaller than the number of normal returns our data is imbalanced. For instance, at the threshold level of 1.5 only about 13\% of the returns are significant with the remaining 87\% considered normal. Imbalanced class distribution may cause bias in the learning model whereby the majority labeled samples are better classified than the minority samples. At the same time the minority class samples are often of greater importance. The traditional measures of performance such accuracy or error rate are not appropriate for imbalanced data as they mask the performance of the classifier on the minority samples. 
Therefore, we use the $F_1$-score which is a balanced metric. It takes into account classifier accuracy on both positive and negative samples. The  $F_1$-score is calculated based on \textit{precision} and \textit{recall} via the equation:

\begin{equation*}
F_1 = 2\cdot \frac{precision\cdot recall}{precision+ recall},
\end{equation*}
where $precision = \frac{tp}{tp + fp}$ and $recall = \frac{tp}{tp+fn}$. In addition to $F_1$-score we report the \textit{recall} results of the experiments. Recall represents the fraction of significant instances that were correctly identified. We note that $F_1$ and recall scores may diverge - as in fact would be the case in our study - which indicates low precision of the model.

To ensure validity of the results, we divide the data into training and testing parts via a 70/30 percent temporal split. The testing set is used only once upon completion of the training phase. We used the scikit-learn implementation of RF, SVC, OLS, SVR, LOF, and RC together with their standard settings. We also employed Keras implementation of NNR and NNC using one hidden layer together with the standard settings.

%-----------------------------------------------------------------------------------------------------------------------------------------------------
%-----------------------------------------------------------------------------------------------------------------------------------------------------
\subsection{Results}
\begin{figure}[h!]
\center
\includegraphics[scale=0.65]{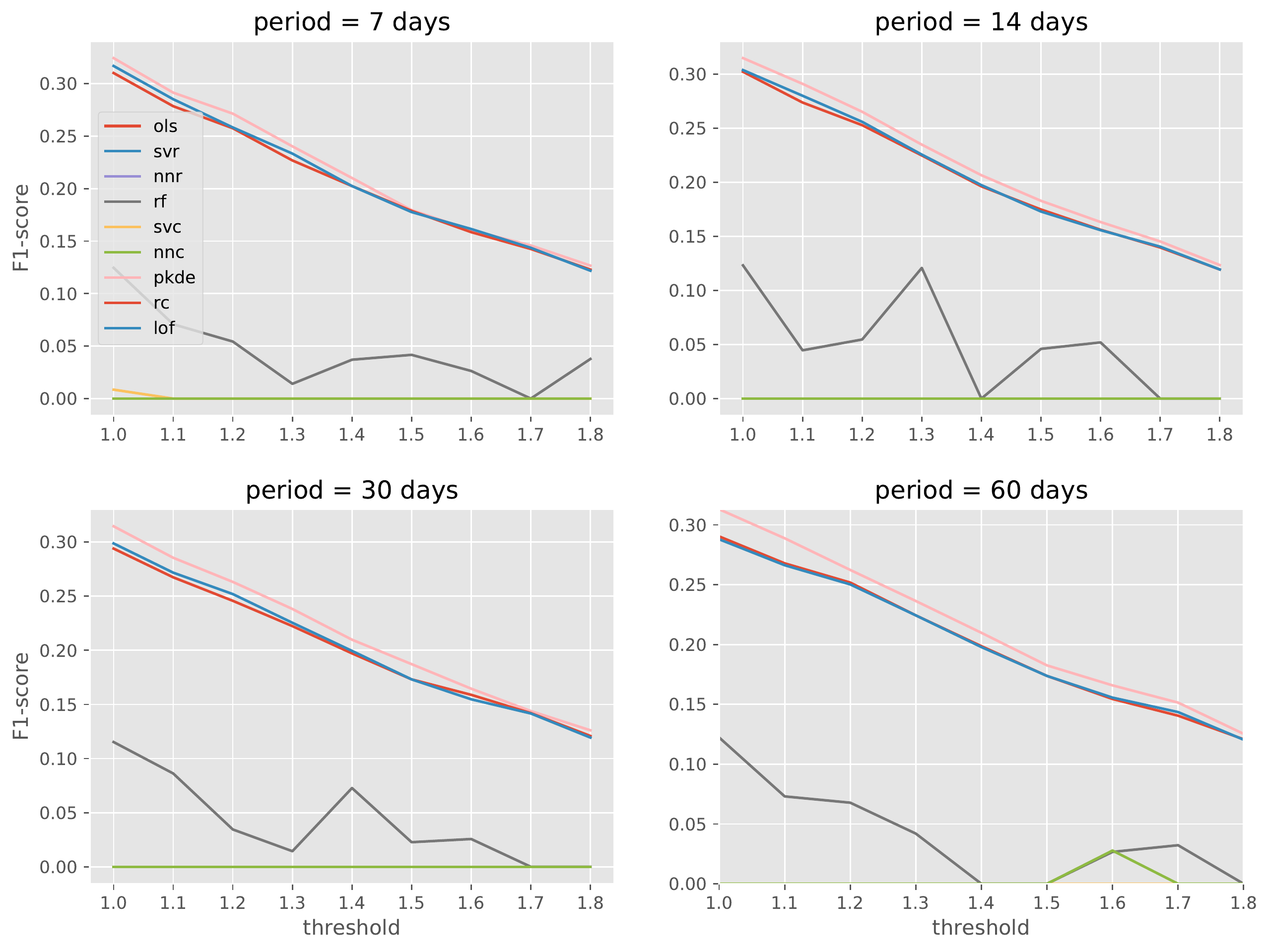}
\caption{$F_1$-scores of learning models for the EUR/USD dataset. The scores are calculated over a range of significance thresholds and based on different number prior days.}
\label{eur_usd}
\end{figure}

\begin{figure}[h!]
\center
\includegraphics[scale=0.65]{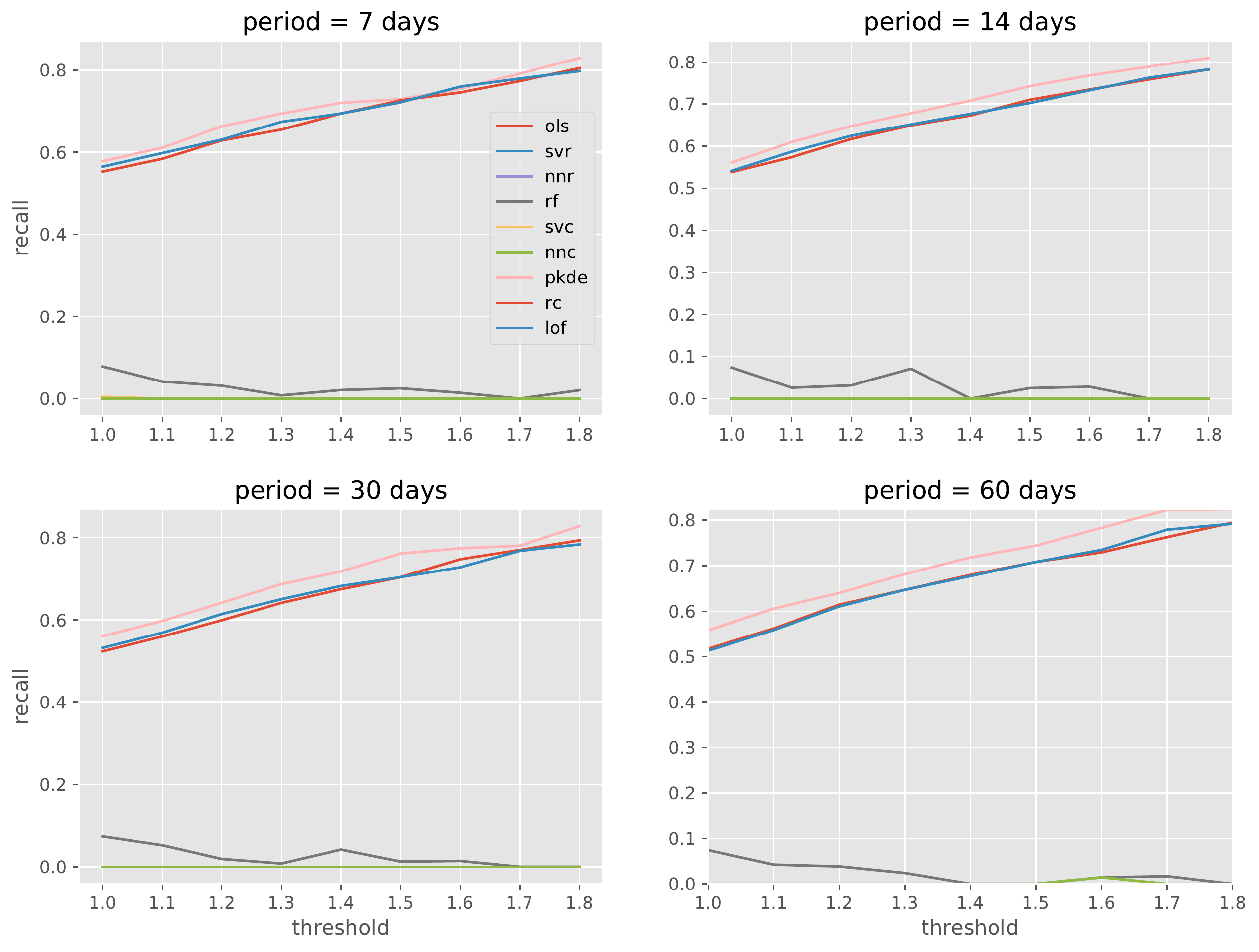}
\caption{\textit{Recall} scores of learning models for the EUR/USD dataset. }
\label{eur_usd_recall}
\end{figure}

We begin our analysis with application of the forecasting methods to the EUR/USD exchange rate data. The graph of EUR/USD daily returns is given in Figure \ref{all_returns}. Our goal is to anticipate significant returns based on prior returns. The $F_1$-scores of forecast tests are presented in Figure \ref{eur_usd}. 
As can be seen from Figure \ref{eur_usd}, the outlier detection methods substantially outperform other models with the PKDE method holding a slight edge over all other methods. In Figure \ref{eur_usd}(a), the models forecast daily return based on the returns over the previous 7 days. The models are tested using different thresholds of significance indicated on the $x$-axis. It can be seen that outlier methods perform consistently well over a range of threshold values. In addition, the number of past days used in prediction does not affect the performance results. We note that classification models perform second best among the tested groups. The best classifier is RF which consistently outperforms NNC and SVC albeit still far behind the outlier detection methods. Unsurprisingly, the regression methods do not perform well. Since the regression models try to estimate the actual daily returns, which are essentially random, they get overwhelmed by the noise. We also observe from Figure \ref{eur_usd} that the $F_1$-score performance of outlier detection methods steadily deteriorates as the threshold level increases. In other, words outlier detection methods are less accurate at discriminating between extreme and non extreme daily returns. This is in line with random walk model that produces a linearly decreasing $F_1$-curve. 
In contrast, the recall scores in Figure \ref{eur_usd_recall} show that outlier detection algorithms improve with increase in threshold level. It indicates that the outlier detection methods are better at identifying the outliers at high thresholds. We note that the results are statistically robust in the sense that a random walk model has a recall of 0.5. Note that the detection methods outperform other models in recall by a substantial margin. PKDE performs the best albeit by a small margin. The results remain consistent across different periods and thresholds. 

\begin{figure}[h!]
\center
\includegraphics[scale=0.65]{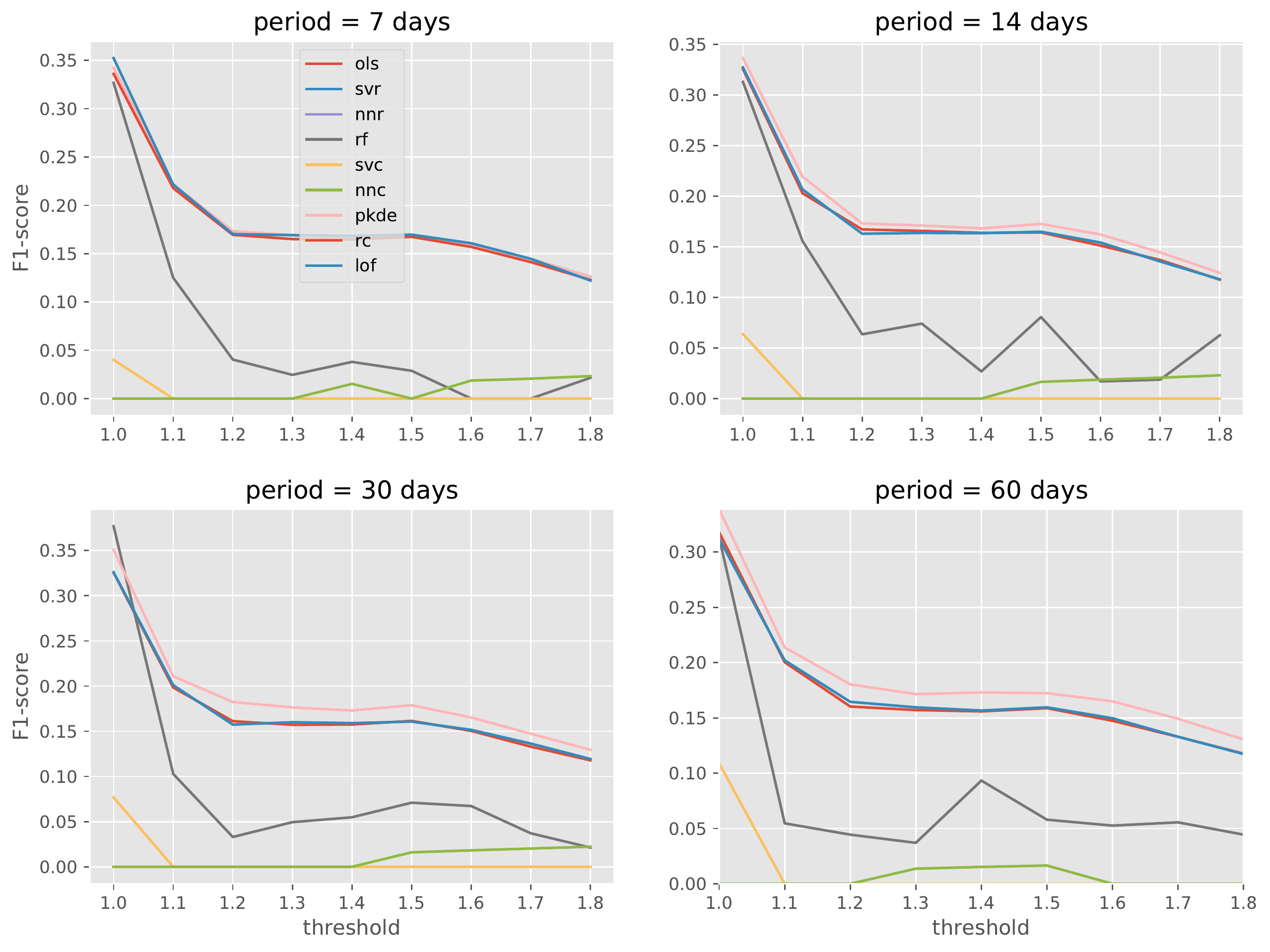}
\caption{$F_1$-scores of learning models for the GBP/USD dataset. The scores are calculated over a range of significance thresholds and based on different number prior days.}
\label{pound_usd}
\end{figure}

Next, we investigate model performance on GBP/USD daily return data (Figure \ref{all_returns}). As can be seen from  Figure \ref{pound_usd}, the forecast performance of classifiers and outlier detection methods has decreased slightly over the EUR/USD data. Interestingly, at the lowest threshold levels the RF model performs as well  as the outlier detection methods. However, at the mid and high threshold levels the outlier detection methods substantially outperform the RF model. The performance of outlier detection methods drops at the beginning and then flattens out. In other words, outlier detection methods perform evenly at forecasting significant and moderately significant return values. All three outlier detection methods produce similar results. However, PKDE produces  better results in the 30 and 60 day periods. The classification models with exception of the RF model perform poorly. Regression models, as expected, do not perform well. The daily return values of GBP/USD exchange rate (Figure \ref{all_returns}) do not have any trend and are evenly distributed around mean zero. Therefore, regression methods will produce models with high bias and low variance which are ineffective at predicting significant values 

\begin{figure}[h!]
\center
\includegraphics[scale=0.65]{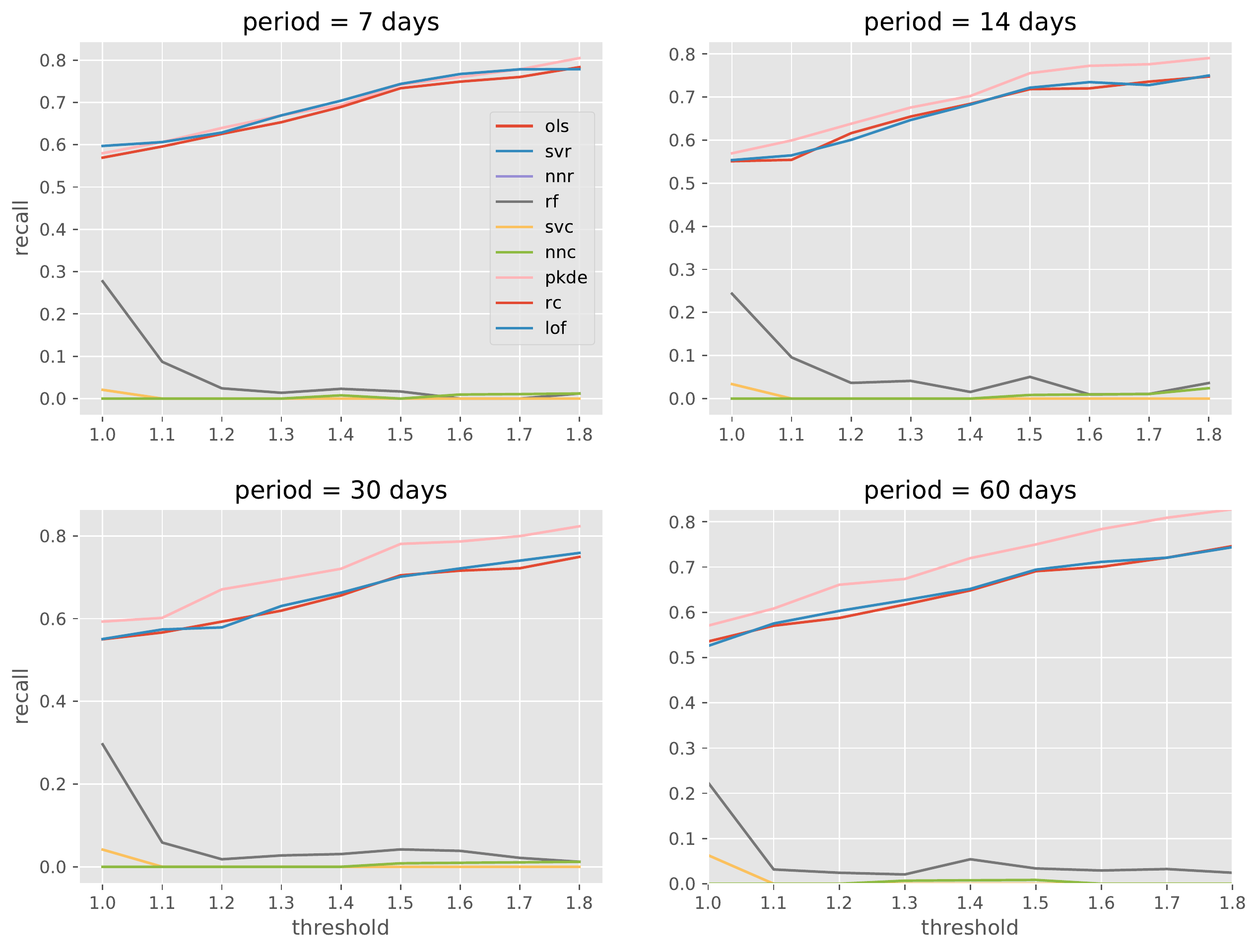}
\caption{\textit{Recall} scores of learning models for the GBP/USD dataset. }
\label{pound_usd_recall}
\end{figure}

The recall results  on GBP/USD daily return data are similar to that of EUR/USD data (Figure \ref{pound_usd_recall}). The outlier detection methods substantially outperform all other models across the different thresholds and periods. Only the RF model manages to achieve modest success at low significance levels. The recall scores  increase with the threshold which indicates that the outlier detection models are effective in identifying a large fraction of outliers at higher significance levels. We note again that PKDE is overall best performer especially in the 30 and 60 day periods.

\begin{figure}[h!]
\center
\includegraphics[scale=0.65]{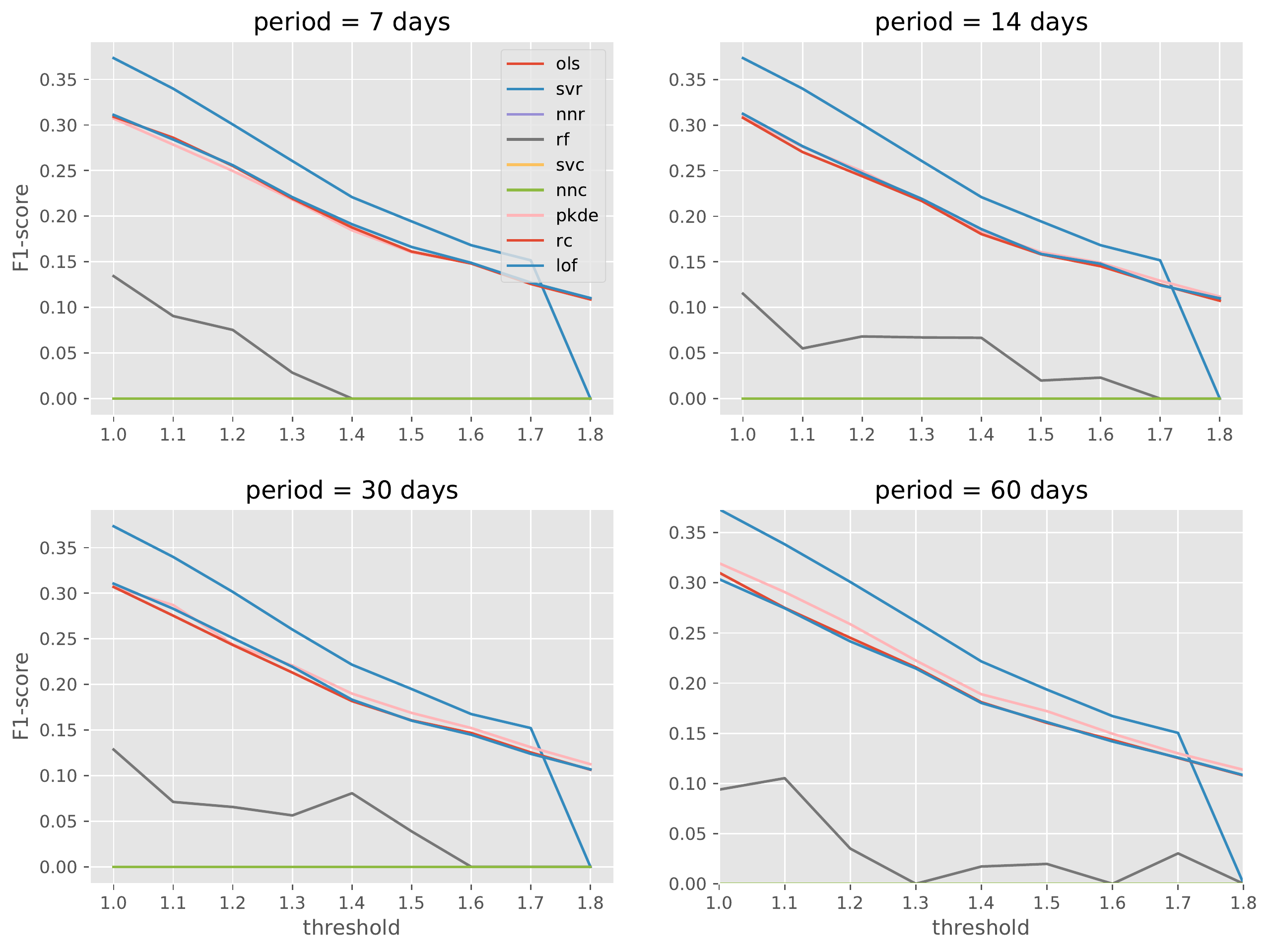}
\caption{$F_1$-scores of learning models for the YEN/USD dataset. The scores are calculated over a range of significance thresholds and based on different number prior days.}
\label{yen_usd}
\end{figure}

Our third forecasting experiment is based on YEN/USD dataset (Figure \ref{all_returns}). As can be seen from Figure \ref{yen_usd}, the outlier detection methods outperform classification and most of the regression methods.
In general, the results of the experiments on YEN/USD are in line with our findings from the previous two experiments. The only noteworthy observation is the performance of the SVR model which produced unusual results. The SVR model is the best overall model, outperforming the outlier detection models, though its score drops to zero in the end. The results of recall scores for the  YEN/USD data are also mostly in line with the previous recall scores for  EUR/USD and  GBP/USD. 

\begin{figure}[h!]
\center
\includegraphics[scale=0.65]{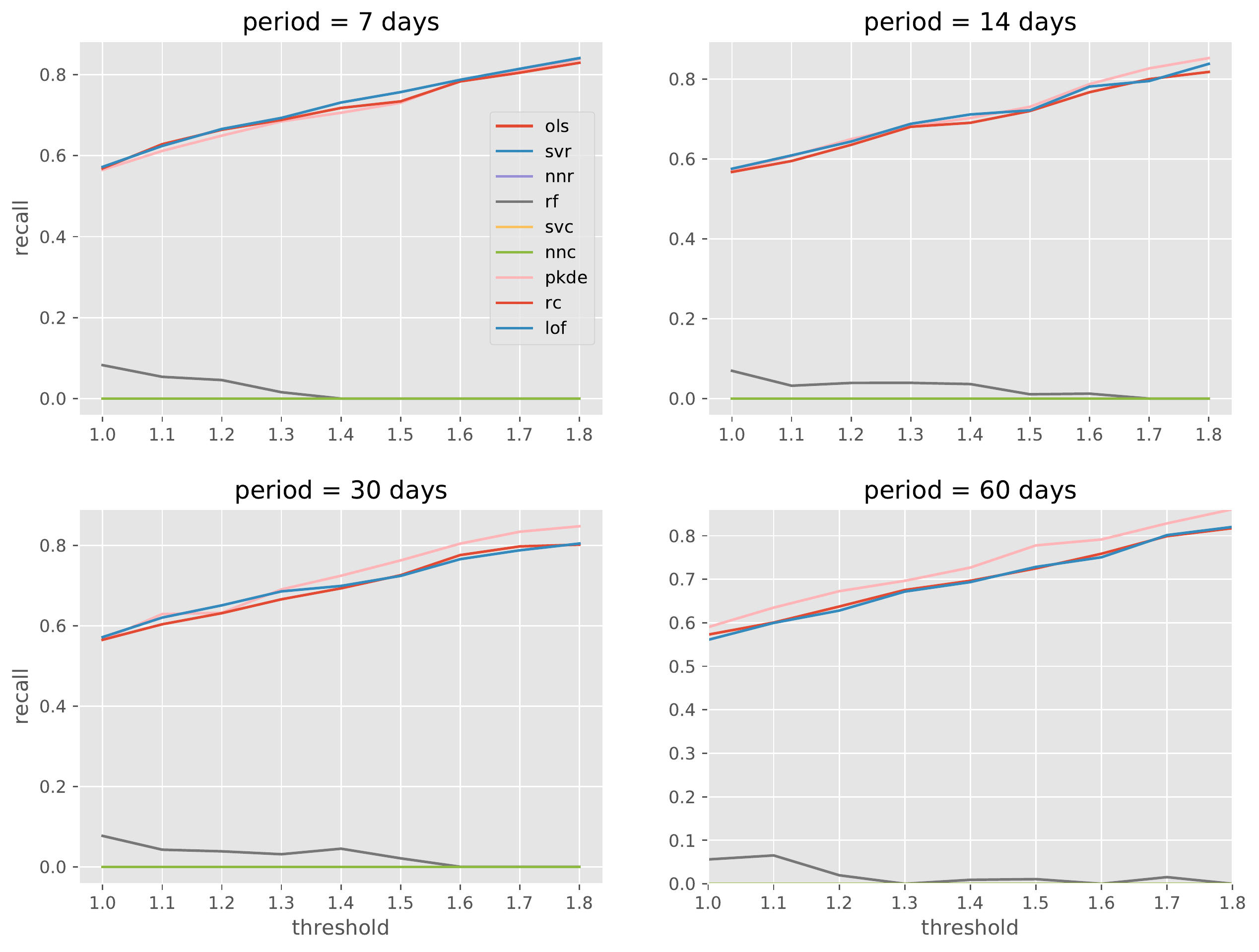}
\caption{$Recall$ scores of learning models for the YEN/USD dataset. }
\label{yen_usd_recall}
\end{figure}

Our final experiment is based on AUD/USD daily returns (Figure \ref{all_returns}). The results of the experiment on AUD/USD data are largely in line with the previous three experiments. In fact, the obtained results are very similar to the EUR/USD based experiment though lower by approximately  5\%. Likewise, the recall results are similar to the previous results though lower by about 10\%.

\begin{figure}[h!]
\center
\includegraphics[scale=0.65]{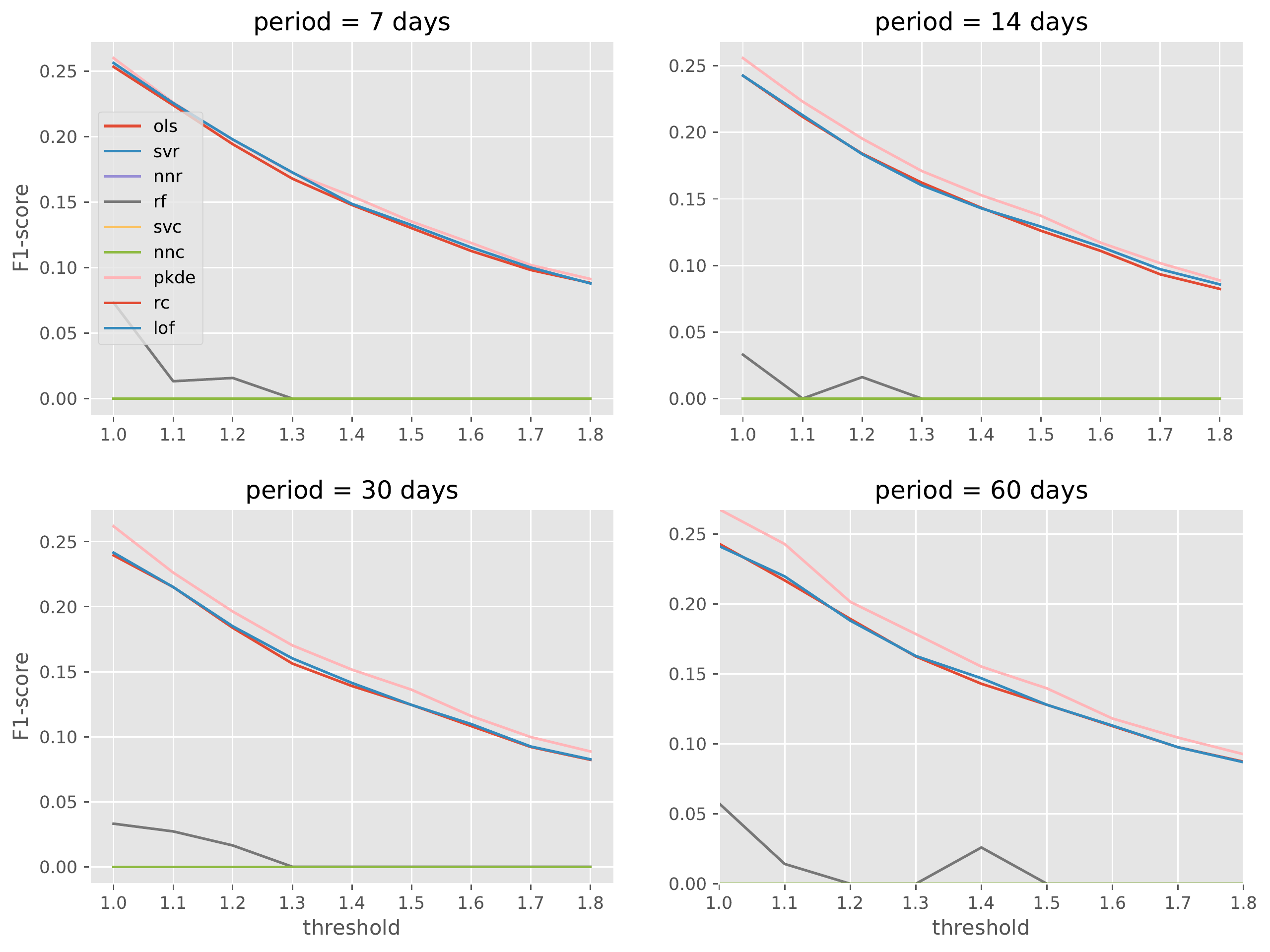}
\caption{$F_1$-scores of learning models for the AUD/USD dataset. The scores are calculated over a range of significance thresholds and based on different number prior days.}
\label{aud_usd}
\end{figure}

\begin{figure}[h!]
\center
\includegraphics[scale=0.65]{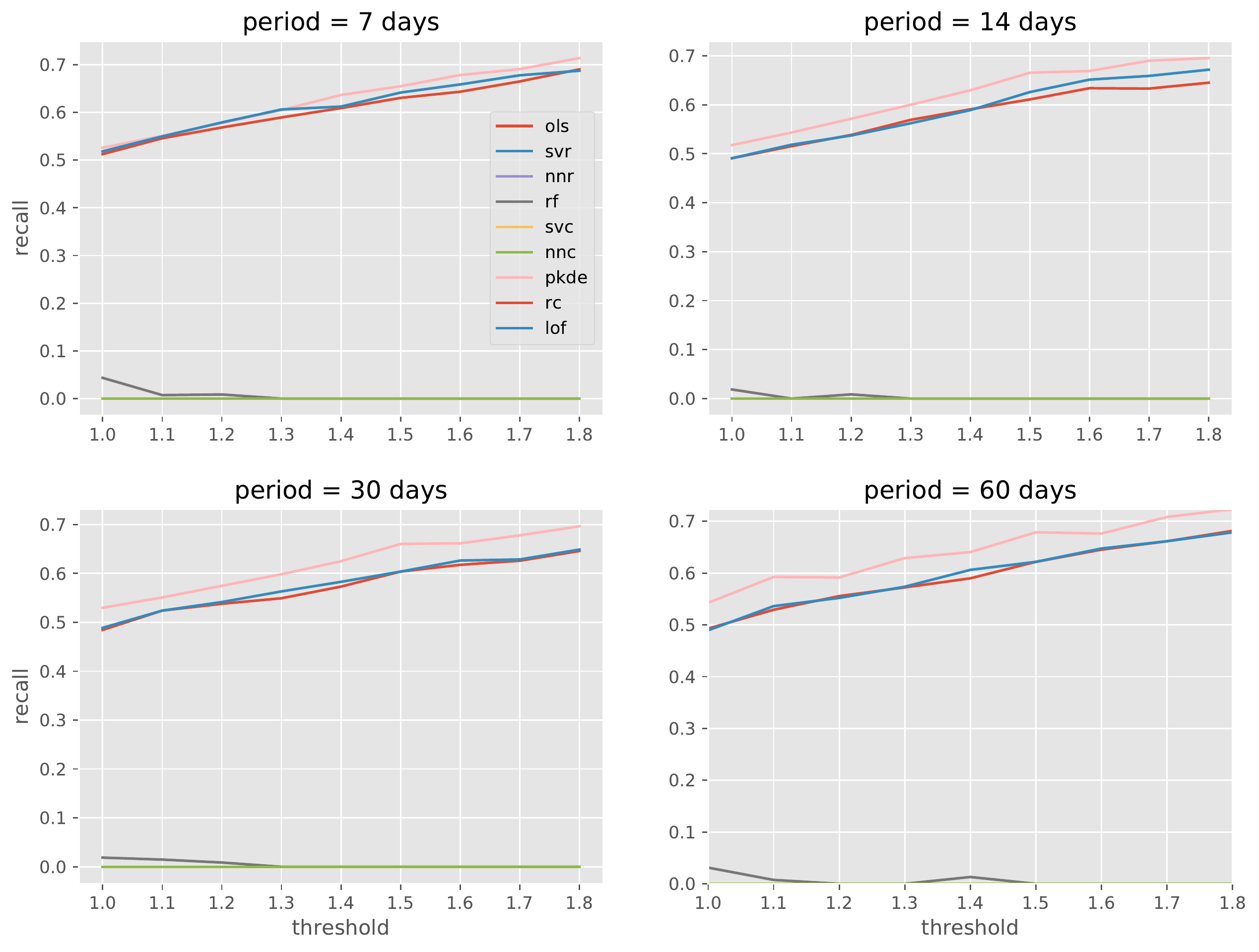}
\caption{$Recall$ scores of learning models for the AUD/USD dataset. }
\label{aud_usd_recall}
\end{figure}

\begin{figure}[h!]
\center
\includegraphics[scale=0.65]{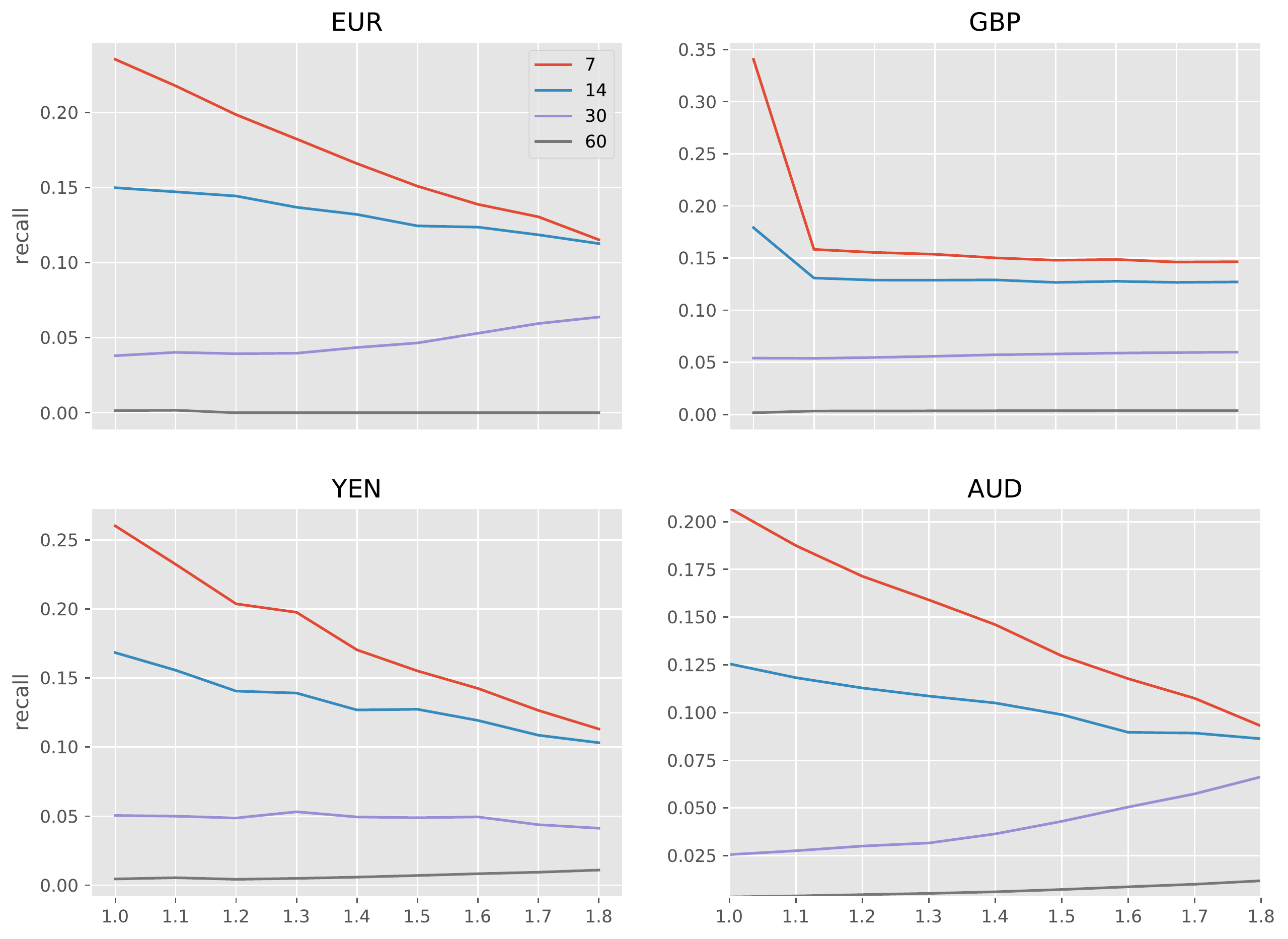}
\caption{$F_1$-scores using the RSI  model for different periods and currencies.}
\label{rsi_plots}
\end{figure}

The results of applying the RSI model to forecast significant daily returns are presented in Figure  \ref{rsi_plots}. The $F_1$-scores are similar across different currency pars with exception of the GBP/USD data. The best forecast model is obtained using the 7-day lookback period. The 60-day lookback model produces the worst results. The 7-day model generally starts with $F_1$-score of 0.25 at the low threshold level and steadily decreases to 0.10 at the highest threshold levels. Although these results are better than most of the classification and the regression models tested in the paper they are still lower than the outlier detection models.

\begin{figure}[h!]
\center
\includegraphics[scale=0.65]{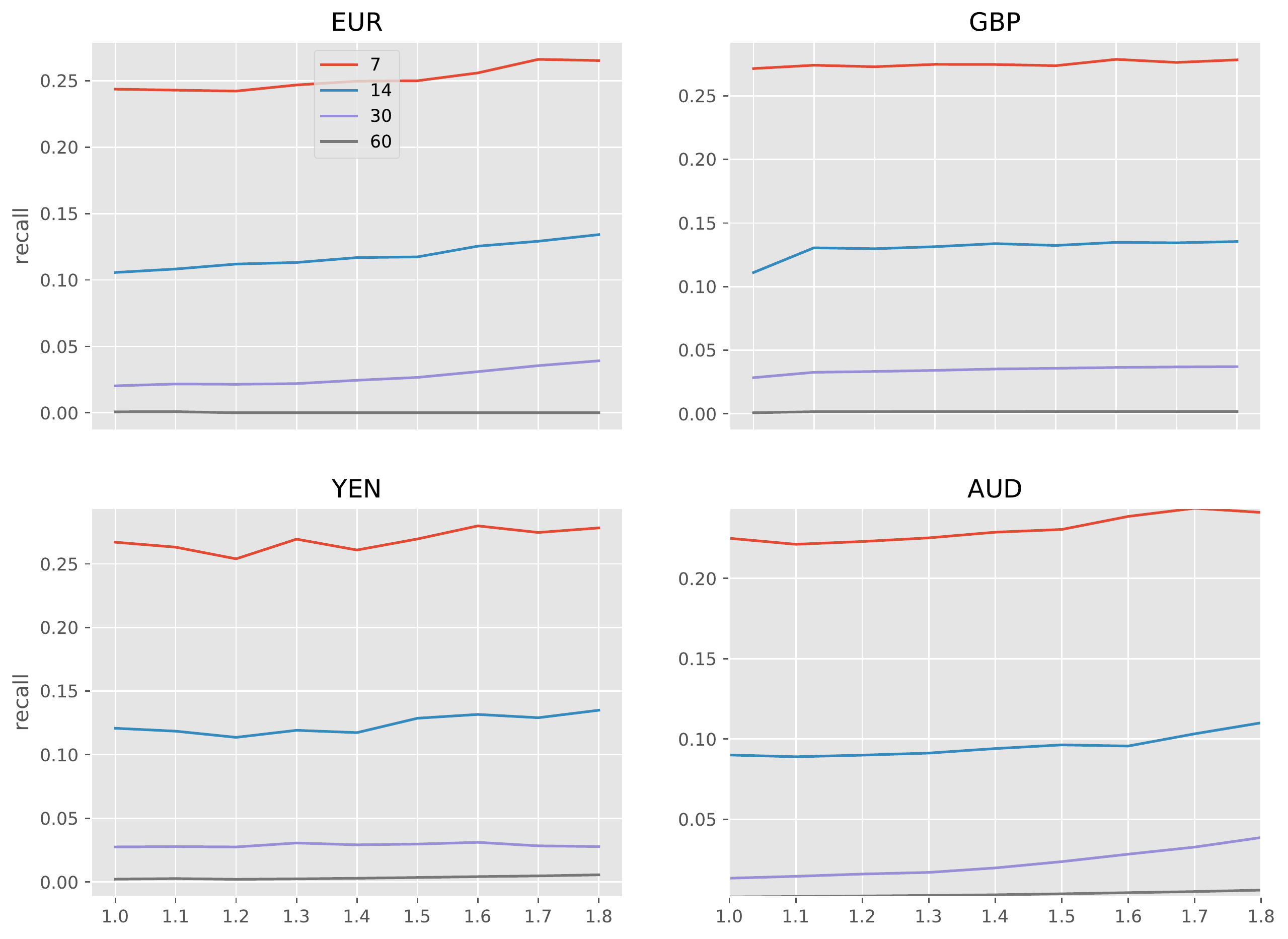}
\caption{$Recall$ scores using the RSI  model for different periods and currencies.}
\label{rsi_recall_plots}
\end{figure}

The recall scores of the RSI model are presented in Figure  \ref{rsi_recall_plots}. We can see that the results are largely similar across different currencies. The best and worst performing models are the 7-day and 60-day models respectively. The performance of all models is even across different threshold levels. We surmise that the information about the total gains and losses across a longer periods is  not  useful, and even misleading. 
Note that the outlier detection models clearly outperform the benchmark RSI models in recall scores by a considerable margin.

%-----------------------------------------------------------------------------------------------------------------------------------------------------%-----------------------------------------------------------------------------------------------------------------------------------------------------
\clearpage
\section{Conclusion}
The existing research deals with forecasting either the actual currency exchange rate or direction of rate change. In this paper, we studied the problem of forecasting significant changes in exchange rate.
To this end, we analyzed a number of machine learning approaches. We examined three major categories of machine learning algorithms (Table \ref{algo_list}): regression, classification, and outlier detection. We performed extensive evaluations using 10-year historical data on four major currency exchange pairs. In addition, we proposed a novel approach to forecasting using outlier detection methods. The results were benchmarked using an RSI forecasting model.  Our findings are summerized below:

\begin{itemize}
\item Outlier detection methods substantially outperform other machine learning approaches in forecasting significant daily returns. In particular, the recently proposed PKDE model produces overall best results among all tested methods especially using 30 and 60-day prior returns.
\item Classification approaches are the second best group among the tested methods with RF outperforming SVR and ANN models.
\item Regression is generally an ineffective approach to forecasting significant returns.
\item The $F_1$-scores deteriorate when the threshold level is increased which indicates that the models struggle to distinguish between significant and insignificant instances at higher levels. In particular, only a small fraction of positively labeled instances are truly positive. In contrast, the recall scores improve when the threshold level is increased. It indicates that the models are better at identifying the significant instances at higher levels.
\item The results are generally consistent across different currency pairs, thresholds, and periods which indicates robustness of the experiments.
\end{itemize}

The outcomes of this study can be applied in other settings where predicting the exact value of a continuous target variable is infeasible and instead one endeavors to tackle a more realistic problem of predicting significant changes in value of the target. In the future, a more in depth study using a larger number of input features would further illuminate the issues and the advantages of predicting significant daily returns.
\newline
%----------------------------------------------------------------------------------------------------------------------------------------------------
%----------------------------------------------------------------------------------------------------------------------------------------------------

\noindent
\textbf{Data Availability Statement}
\\
\noindent The data that support the findings of this study are publicly available from Macrotrends LLC at  https://www.macrotrends.net/charts/exchange-rates \cite{macrotrends}

%-----------------------------------------------------------------------------------------------------------------------------------------------------%-----------------------------------------------------------------------------------------------------------------------------------------------------

\end{document}